\documentclass{article}

\PassOptionsToPackage{numbers, compress}{natbib}

\usepackage[preprint]{neurips_2026}

\usepackage[utf8]{inputenc} % allow utf-8 input
\usepackage[T1]{fontenc}    % use 8-bit T1 fonts
\usepackage{hyperref}       % hyperlinks
\usepackage{url}            % simple URL typesetting
\usepackage{booktabs}       % professional-quality tables
\usepackage{amsfonts}       % blackboard math symbols
\usepackage{amssymb}        % additional math symbols
\usepackage{amsmath}        % math environments
\usepackage{nicefrac}       % compact symbols for 1/2, etc.
\usepackage{microtype}      % microtypography
\usepackage{xcolor}         % colors

\usepackage{multirow}
\usepackage{float}
\usepackage{graphicx}
\usepackage{enumitem}
\setlist[itemize]{noitemsep}

\title{Unlocking High-Fidelity Molecular Generation from Mass Spectra via Dual-Stream Line Graph Diffusion}

\author{
  Xujun Che$^{1,3}$\\
  \texttt{xche@charlotte.edu} \\
  \And
  Xiuxia Du$^{2,3}$ \\
  \texttt{xdu4@charlotte.edu} \\
  \AND
  Depeng Xu$^{1,3}$ \\
  \texttt{dxu7@charlotte.edu} \\
  \\
  $^{1}$ Department of Software and Information Systems,  \\University of North Carolina at Charlotte, Charlotte, NC 28223\\
  \\
  $^{2}$ Department of Bioinformatics and Genomics, \\
  University of North Carolina at Charlotte, Charlotte, NC 28223\\
  \\
  $^{3}$ Center for Environmental Monitoring and Informatics Technologies for Public Health \\
  University of North Carolina at Charlotte, Charlotte, NC 28223
}

\begin{document}

\maketitle

\begin{abstract}
De novo molecular generation from tandem mass spectra is a challenging inverse problem whose core difficulty lies in the circular dependency between atom-level and bond-level reasoning: determining a bond's type requires knowing its endpoint atoms' chemical environment, yet an atom's environment is in turn defined by its incident bonds.
Existing graph diffusion methods process atoms and bonds within a single computation stream, where atom-bond information synchronization can only occur implicitly across layers.
We argue that this single-stream paradigm, rather than the choice of any particular aggregation kernel, is a key architectural bottleneck.
We propose DualLGD (Dual-stream Line Graph Diffusion), which reformulates molecular graph denoising as the alternating solution of two coupled subproblems: atom-level reasoning and bond-level reasoning, each operating in its own dedicated representation space.
The line graph provides a natural mathematical construction for the bond space, in which bond angles, dihedrals, conjugation chains, and rings correspond to local topological motifs between bonds.
Incidence-constrained bidirectional cross-attention synchronizes the two streams at every layer, ensuring that each atom attends only to its incident bonds and vice versa, respecting the fundamental chemical principle that an atom's environment is determined by its bonding context.
On the NPLIB1 and MassSpecGym benchmarks, DualLGD achieves top-1 accuracy of 34.37\% and 23.89\%, approximately $3\times$ the previous state of the art. 
Ablation studies confirm the architecture as the primary source of improvement: DualLGD without any pre-training already surpasses the previous best fully pretrained model. 
Code and pretrained models are publicly available at \url{https://github.com/du-lab-data-science/DualLGD}.
\end{abstract}

\section{Introduction}
\label{sec:introduction}

Tandem mass spectrometry (MS/MS) is a cornerstone analytical technique in metabolomics, drug discovery, and environmental chemistry, providing characteristic fragmentation patterns that encode structural information about unknown molecules~\cite{wang2016sharing}.
However, the vast majority of experimentally acquired spectra remain unidentified, as spectral library search is fundamentally limited to previously characterized compounds~\cite{horai2010massbank,duhrkop2019sirius,allen2015competitive,duhrkop2015searching,ruttkies2016metfrag,wang2021cfm}.
De novo molecular structure generation from mass spectra, the task of inferring a complete molecular graph given only a spectrum and molecular formula~\cite{che2026comparative,goldman2023mist,duhrkop2014molecular}, offers a principled route to identifying novel molecules, yet remains an extremely challenging inverse problem~\cite{schymanski2017critical}: a single spectrum is consistent with a combinatorially large space of candidate structures.

Recent years have witnessed rapid progress on this problem.
Early approaches generate molecular structures as SMILES strings~\cite{weininger1988smiles,spec2mol, msbart}, but the sequential nature of SMILES imposes an arbitrary atom ordering that fails to reflect the permutation-invariant structure of molecular graphs.
Scaffold-based methods~\cite{madgen, msanchor} decompose the problem into substructure retrieval and completion, but their coverage is bounded by the scaffold vocabulary.
Graph diffusion methods have emerged as a more principled alternative: DiffMS~\cite{diffms} first introduced discrete graph diffusion for spectrum-conditioned molecular generation and established a fingerprint-conditioned pretraining paradigm, and MBGen~\cite{mbgen} subsequently improved performance by incorporating many-body attention to model bond-bond interactions. However, both remain single-stream architectures in which bonds are maintained as edge attributes rather than first-class representation objects.

A fundamental challenge in molecular graph denoising is the \emph{circular dependency} between atom-level reasoning and bond-level reasoning.
Determining a bond's type (single, double, triple, or aromatic) requires knowledge of its endpoint atoms' chemical environment (hybridization, electronegativity, steric context), yet an atom's chemical environment is in turn defined by the collection of bonds it participates in.
Resolving this circular dependency demands that atom-level and bond-level inference be tightly coupled and iteratively refined.
Existing graph diffusion methods, despite differing in the strength of their edge aggregation kernels, share a common structural limitation: they process atoms and bonds within a \emph{single computation stream}.
In DiffMS~\cite{diffms}, bond information resides in edge attributes whose dimensionality, update rules, and computation paths are subordinate to the node-level computation.
MBGen~\cite{mbgen} enriches edge updates through many-body attention~\cite{pmlr-v235-hussain24a}, yet the bonds still lack an independent representation space: each bond $(i,j)$ can only interact with bonds sharing an endpoint atom, and atom-bond synchronization occurs only implicitly across layers.
This is not a limitation of any specific method, but a \emph{bottleneck of the single-stream architectural paradigm} itself: after the atom representations are updated at layer $l$, the bond representations still see the stale atom states from layer $l-1$, and vice versa.

In this work, we propose \textbf{DualLGD} (\textbf{Dual}-stream \textbf{L}ine \textbf{G}raph \textbf{D}iffusion), which reformulates molecular graph denoising as the alternating solution of two coupled subproblems: atom-level reasoning and bond-level reasoning, each operating in a dedicated representation space.
The core idea is to promote bonds from secondary edge attributes to \emph{first-class nodes} in a line graph, granting each bond its own representation vector and an independent computation path.
In this bond representation space, chemically meaningful substructures such as bond angles, dihedral angles, conjugation chains, and ring systems correspond to simple topological patterns between bond nodes (edges, short paths, and cycles), and global self-attention further allows any two bonds to interact directly within a single layer regardless of whether they share an endpoint atom.
At every layer, the two streams are explicitly synchronized through bidirectional cross-attention constrained by the molecular incidence matrix, ensuring that each atom attends only to its incident bonds and each bond attends only to its endpoint atoms, respecting the chemical principle that an atom's environment is determined solely by its bonding context.
We evaluate DualLGD on two established benchmarks for de novo molecular generation from mass spectra: NPLIB1~\cite{nplib1} and MassSpecGym~\cite{massspecgym}.
DualLGD achieves top-1 accuracy of 34.37\% and 23.89\%, respectively, representing approximately $3\times$ improvement over the previous state of the art on both benchmarks. Ablation studies reveal that DualLGD without any pre-training (25.53\%) already surpasses the previous best fully pretrained result (12.20\%), demonstrating that the performance leap originates from the dual-stream architectural paradigm rather than the training recipe.
While developed for molecular generation, the dual-stream line graph paradigm is applicable to any graph diffusion setting where edge semantics play a decisive role.

Our contributions are as follows:
\begin{itemize}
    \item We propose DualLGD, which reformulates molecular graph denoising as the alternating solution of atom-level and bond-level subproblems. By promoting bonds to first-class nodes in a line graph, each bond gains a dedicated representation vector and an independent computation path. Chemically meaningful substructures (bond angles, dihedrals, conjugation chains, ring systems) correspond to local topological patterns between bond nodes, and global self-attention over this bond space lets any two bonds interact directly within a single layer, providing a strong inductive bias for bond-level reasoning.
    \item We introduce incidence-constrained bidirectional cross-attention that synchronizes the atom and bond streams at every layer, where each atom attends only to its incident bonds and each bond attends only to its endpoint atoms. This per-layer, topology-aware synchronization overcomes the implicit, cross-layer synchronization bottleneck of single-stream architectures, directly addressing the circular dependency between atom-level and bond-level inference.
    \item We achieve approximately $3\times$ improvement in top-1 accuracy over the previous state of the art on both NPLIB1 and MassSpecGym. Ablation studies confirm that the dual-stream architecture is the primary source of improvement, with DualLGD surpassing the previous state of the art even without any pre-training.
\end{itemize}

\section{Related Work}
\label{app:related}

\paragraph{De novo molecular generation from mass spectra.}
Sequence-based methods cast spectrum-to-structure prediction as a translation task. MSNovelist~\cite{msnovelist} chains fingerprint prediction via CSI:FingerID~\cite{duhrkop2015searching} with LSTM-based SMILES~\cite{weininger1988smiles} decoding. Spec2Mol~\cite{spec2mol} couples a CNN spectrum encoder with a pretrained SMILES decoder, while MS-BART~\cite{msbart} unifies spectra and SELFIES~\cite{krenn2020self} in a shared tokenized space. A recurring drawback of these approaches is that string representations serialize an inherently unordered graph, making it difficult to enforce hard chemical constraints such as a fixed molecular formula. Scaffold-based methods sidestep full graph generation by first retrieving molecular substructures and then completing them: MADGEN~\cite{madgen} retrieves Murcko~\cite{bemis1996properties} scaffolds conditioned on spectral features and completes the remaining structure through a conditional diffusion process, while MSAnchor~\cite{msanchor} grows anchor fragments guided by spectral attention. Their accuracy, however, is bounded by the coverage and quality of the scaffold vocabulary. 

More recently, graph diffusion methods have gained traction. DiffMS~\cite{diffms} introduced discrete graph diffusion~\cite{digress} for this task and proposed large-scale fingerprint-conditioned pretraining of the decoder. MBGen~\cite{mbgen} built on this framework by incorporating many-body attention~\cite{pmlr-v235-hussain24a} to capture higher-order bond-bond interactions. A structural limitation shared by both approaches is that bonds are represented as edge attributes that lack an independent representation space: edge states are updated through edge-modulated node attention layers, and consistency between atom and bond representations is established only indirectly through successive layers. This observation motivates the dual-stream design of DualLGD.

\paragraph{Discrete graph diffusion.}
Denoising diffusion probabilistic models~\cite{ho2020denoising}, originally developed for continuous data, have been extended to discrete state-spaces through structured categorical transition matrices~\cite{austin2021structured}. DiGress~\cite{digress} adapts this framework to graph generation by defining a discrete diffusion process that jointly corrupts and denoises categorical node and edge attributes, using a graph transformer as the denoising network with marginal-preserving noise schedules. DiffMS~\cite{diffms} and MBGen~\cite{mbgen} build upon DiGress for spectrum-conditioned molecular generation, inheriting its cosine noise schedule, marginal transition matrices, and cross-entropy training objective. DualLGD likewise inherits this diffusion framework; our contribution is orthogonal and concerns the denoising network architecture: we decouple atom-level and bond-level reasoning into two synchronized streams, rather than modifying the diffusion process itself.

\paragraph{Line graphs in molecular representation learning.}
Promoting bonds to first-class objects in graph neural networks has proven valuable for molecular property prediction. DimeNet~\cite{Gasteiger2020Directional} pioneered directional message passing, where messages are associated with directed edges and modulated by bond angles, an operation conceptually equivalent to convolution on the line graph. GemNet~\cite{gasteiger2021gemnet} extended this idea to two-hop edge interactions that incorporate dihedral angles, corresponding to length-two paths in the line graph. ALIGNN~\cite{choudhary2021atomistic} makes the line graph explicit, alternating edge-gated convolution on the bond graph and its line graph so that pair and angular interactions are captured in separate passes. GEM~\cite{fang2022geometry} follows a similar strategy, constructing a bond-angle graph (the line graph of the molecular graph) alongside the atom-bond graph for concurrent message passing. These architectures are, however, limited to discriminative settings where the molecular topology is known and fixed. In graph denoising for molecular generation, the topology itself is uncertain and must be recovered jointly with atom and bond types, giving rise to mutual dependence between the two inference subproblems that demands explicit synchronization at every denoising layer. DualLGD brings the line graph into the generative diffusion setting and addresses this coupling through incidence-constrained bidirectional cross-attention, providing layer-wise, structurally grounded communication between the atom and bond streams.

\section{Methodology}
\label{sec:methodology}

\begin{figure}[t]
    \centering
    \includegraphics[width=\textwidth]{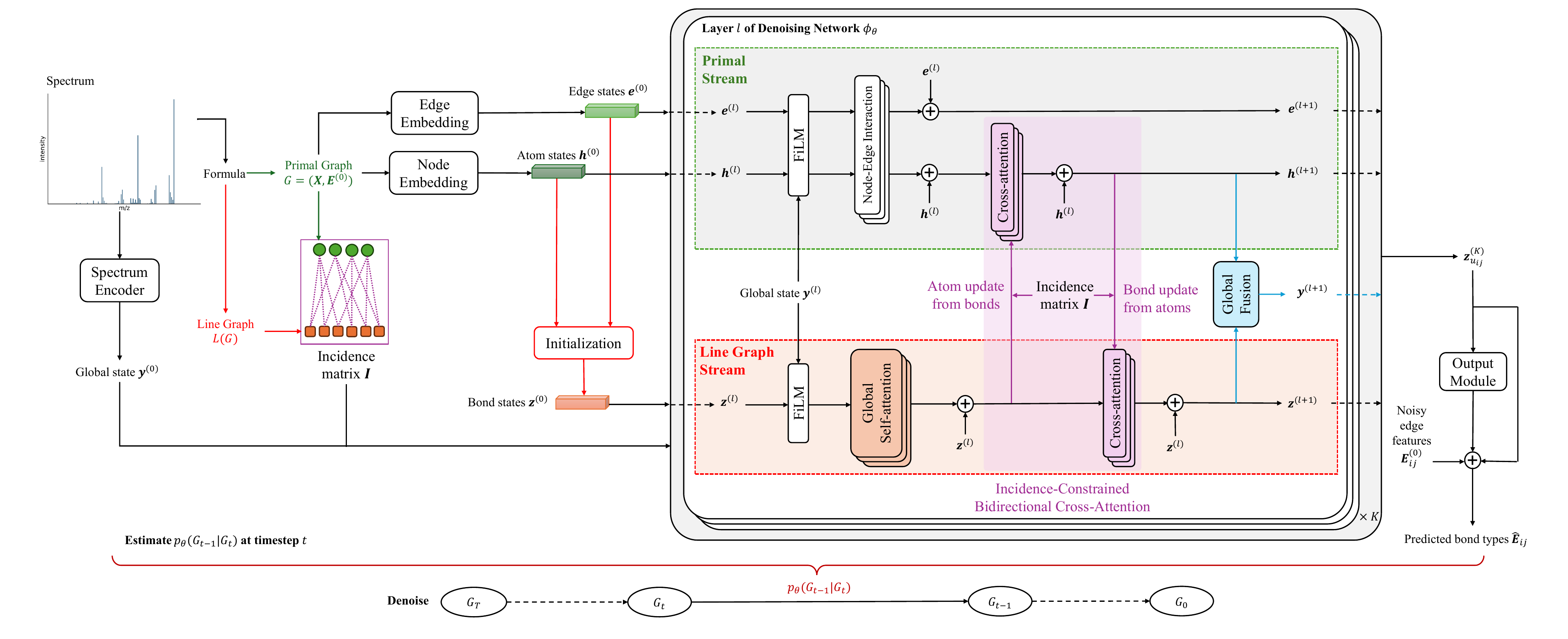}
    \caption{
    % Overview of the DualLGD denoising network. Each layer applies FiLM conditioning, primal stream update, line graph stream update, incidence-constrained cross-attention, and global fusion. The primal stream operates on atom-level representations while the line graph stream enables global bond-bond attention in a decoupled representation space. The two streams exchange information through bidirectional cross-attention masked by the incidence matrix.
    Overview of the DualLGD denoising network. The primal graph stream and line graph stream perform atom- and bond-level reasoning, respectively, and exchange information through incidence-constrained bidirectional cross-attention at each layer.
    }
    \label{fig:architecture}
\end{figure}

 We present DualLGD, a dual-stream diffusion framework that reformulates molecular graph denoising as the alternating solution of two coupled subproblems: atom-level reasoning in a \emph{primal stream} and bond-level reasoning in a dedicated \emph{line graph stream}.
 The key idea is to promote bonds from secondary edge attributes to \emph{first-class nodes} in a line graph, granting each bond its own representation vector and an independent computation path.
 At every layer, the two streams independently update their respective representations, then explicitly synchronize through bidirectional cross-attention constrained by the molecular incidence matrix.
 This per-layer, topology-aware synchronization directly addresses the circular dependency between atom-level and bond-level inference that single-stream architectures can only resolve implicitly across layers.

 \subsection{Problem Formulation}
 \label{sec:problem}

Given a tandem mass spectrum and a molecular formula, the task is to generate a molecular graph $G = (\mathbf{X}, \mathbf{E})$ that is consistent with them.
Here $\mathbf{X} \in \{1, \dots, a\}^{N}$ assigns one of $a$ atom types to each of $N$ heavy atoms, and $\mathbf{E} \in \{0, 1, \dots, b\}^{N \times N}$ is a symmetric matrix of discrete bond-type indices, where $0$ denotes the absence of a bond and $1, \dots, b$ correspond to single, double, triple, and aromatic bonds.
In the denoising network, each discrete entry \(\mathbf{E}_{ij}\) is embedded
from its one-hot encoding in \(\{0,1\}^{b+1}\), and the model outputs logits
over the same \(b+1\) bond categories.
A global conditioning vector $\mathbf{y} \in \mathbb{R}^{d_y}$ is produced by the spectrum encoder. 
% We seek to learn the conditional distribution $p_\theta(G \mid \mathcal{S}, \mathcal{F})$.
% \subsection{Background: Discrete Graph Diffusion}
% \label{sec:diffusion}
Following DiffMS~\cite{diffms} and DiGress~\cite{digress}, we adopt the discrete denoising diffusion framework.
A forward process gradually corrupts the molecular graph $G_0$ over $T$ timesteps via a Markov chain with categorical transition matrices under a cosine noise schedule.
A denoising network $\phi_\theta$ is trained to predict the clean graph from the noisy graph $G_t$ and the conditioning $\mathbf{y}$.
% \begin{equation}
%     \hat{G}_0 = \phi_\theta(G_t,\, t,\, \mathbf{y}),
%     \label{eq:denoise}
% \end{equation}
% from which the posterior $p_\theta(G_{t-1} \mid G_t)$ is computed analytically.
Since the atom types are determined by the known molecular formula, noise is applied only to the bond types $\mathbf{E}$, while $\mathbf{X}$ remains fixed throughout diffusion.
% The training objective is the cross-entropy loss on bond type predictions:
% \begin{equation}
%     L = \mathbb{E}_{t, G_0, G_t}\!\left[-\sum_{i<j} \log \hat{p}_\theta\!\left(\mathbf{E}_{ij} = \mathbf{E}^{0}_{ij} \mid G_t, t, \mathbf{y}\right)\right].
%     \label{eq:loss}
% \end{equation}
During inference, the posterior $p_\theta(G_{t-1} \mid G_t)$ is computed analytically to sample a discrete  $G_{t-1}$ from the predicted distribution.

\subsection{DualLGD Architecture Overview}
\label{sec:overview}

The overall framework consists of a spectrum encoder that maps the spectrum to the conditioning vector $\mathbf{y}$, the DualLGD denoising network $\phi_\theta$, and the diffusion sampling loop.
The spectrum encoder and sampling loop follow DiffMS; our contribution is the denoising network, illustrated in Figure~\ref{fig:architecture}.

The core design principle is to decompose molecular graph denoising into atom-level and bond-level reasoning that alternate within each layer.
DualLGD operates two coupled representation streams.
The \emph{primal stream}, inherited from DiffMS, maintains atom states $\mathbf{h}^{(l)} \in \mathbb{R}^{N \times d_x}$, edge states $\mathbf{e}^{(l)} \in \mathbb{R}^{N \times N \times d_e}$, and a global state $\mathbf{y}^{(l)} \in \mathbb{R}^{d_y}$, updated at each layer $l$ via edge-modulated multi-head self-attention with $n_h$ heads.
The \emph{line graph stream}, introduced in this work, promotes every potential bond to a first-class node and maintains bond states $\mathbf{z}^{(l)} \in \mathbb{R}^{M \times d_e}$ in a dedicated representation space, where $M = \binom{N}{2}$ indexes all atom pairs $(i,j)$ with $i < j$.
At each layer $l$, the two streams first independently update their respective representations (atom-level and bond-level reasoning), then explicitly synchronize through bidirectional cross-attention constrained by the molecular incidence structure, ensuring that atoms attend only to their incident bonds and vice versa.
A global fusion step aggregates information from both streams into $\mathbf{y}^{(l)}$, which in turn modulates both streams at the next layer via FiLM conditioning.
After the final layer $K$, the bond states $\mathbf{z}^{(K)}$ are decoded into bond type predictions.

\begin{figure}[t]
    \centering
    \includegraphics[width=\textwidth]{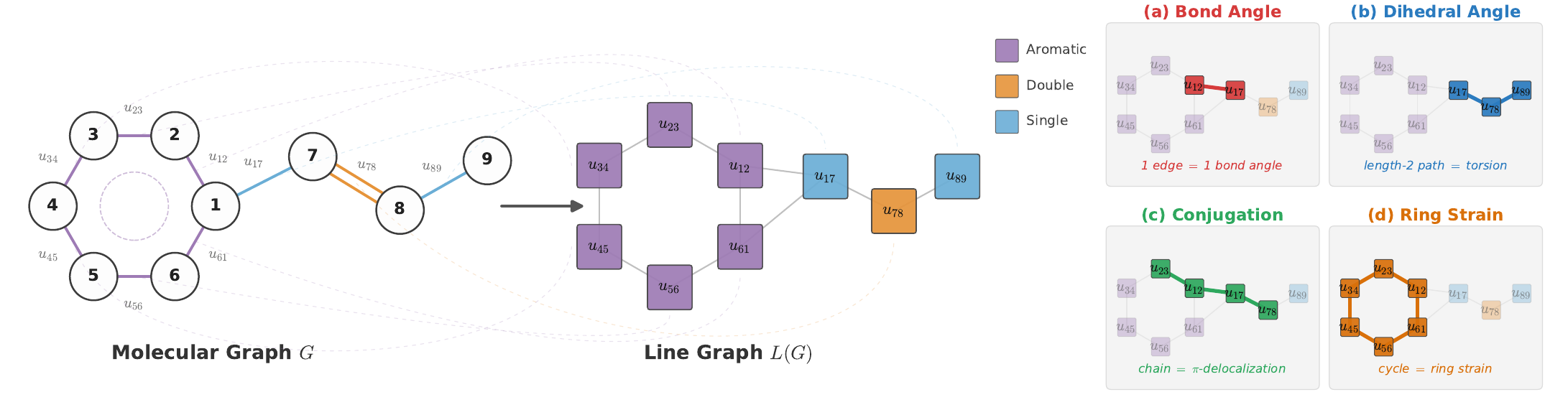}
    \caption{Line graph construction, illustrated on \emph{trans}-propenylbenzene.
    (only heavy atoms are shown, as the molecular graph operates over heavy atoms). 
    Left: the molecular graph $G$ with aromatic (purple), double (orange), and single (blue) bonds. Center: the corresponding line graph $L(G)$, where each bond becomes an independent node and two nodes are adjacent if they share an endpoint atom. Right: four chemical relationships naturally encoded in the line graph topology: (a)~an edge corresponds to a bond angle, (b)~a length-2 path corresponds to a dihedral (torsion) angle, (c)~a chain captures $\pi$-electron delocalization in conjugated systems, and (d)~a cycle reflects ring strain.
    }
    \label{fig:line_graph}
\end{figure}

\subsection{Line Graph Stream: Bonds as First-Class Nodes}
\label{sec:line_graph}

\paragraph{Line graph construction.}
\label{sec:lg_construction}
We construct a line graph $L(G)$ in which every potential bond becomes an independent node in a dedicated representation space (Figure~\ref{fig:line_graph}).
Formally, we enumerate all $M = \binom{N}{2}$ atom pairs $(i, j)$ with $i < j$, each corresponding to a line graph node $u_{ij}$.
Two line graph nodes $u_{ij}$ and $u_{kl}$ are adjacent if and only if they share an atom endpoint $
    \mathbf{L}_{u_{ij}, u_{kl}} = \mathbf{1}\!\left[\{i,j\} \cap \{k,l\} \neq \emptyset\right]
$,
where $\mathbf{L} \in \{0,1\}^{M \times M}$ is the line graph adjacency matrix.
This construction yields topological motifs in the line graph that correspond to chemically meaningful bond-level substructures:
\begin{itemize}[leftmargin=*]
    \item \textbf{Bond-angle motif.} An edge in the line graph corresponds to a bond angle at the shared atom, whose value reflects the atom's hybridization (e.g., ${\sim}109.5^\circ$ for sp$^3$, ${\sim}120^\circ$ for sp$^2$, ${\sim}180^\circ$ for sp).
    \item \textbf{Dihedral motif.} A length-2 path in the line graph corresponds to a dihedral (torsion) angle, which governs molecular conformation and rotational energy barriers.
    \item \textbf{Conjugation motif.} A chain of adjacent line graph nodes provides the substrate on which alternating single/double bond-type patterns indicative of $\pi$-electron delocalization in conjugated and aromatic systems emerge.
    \item \textbf{Ring motif.} A cycle in the line graph corresponds to a ring in the molecule; short cycles in particular indicate small-ring systems whose constrained geometry influences bond-type assignment.
\end{itemize}
By promoting bonds to first-class nodes with their own representation vectors and computation path, DualLGD provides a dedicated substrate on which these motifs can be reasoned about directly, rather than being mediated through edge attributes subordinated to atom-level computation as in single-stream architectures.

\paragraph{Node initialization.}
\label{sec:lg_init}
Each line graph node $u_{ij}$ is initialized by fusing the corresponding edge representation with both endpoint atom representations:
\begin{equation}
    \mathbf{z}^{(0)}_{u_{ij}} = f_{\mathrm{init}}\!\left(\left[\mathbf{e}^{(0)}_{ij} \,\|\, \mathbf{h}^{(0)}_i \,\|\, \mathbf{h}^{(0)}_j\right]\right) \in \mathbb{R}^{d_e},
    \label{eq:lg_init}
\end{equation}
where $[\cdot \| \cdot]$ denotes concatenation and $f_{\mathrm{init}}: \mathbb{R}^{d_e + 2d_x} \to \mathbb{R}^{d_e}$ is a two-layer network with GELU activation.
This provides each bond with an initial representation that captures both its current (noisy) type and the chemical context of its endpoint atoms.

\paragraph{Bond-level reasoning via global self-attention.}
\label{sec:lg_stream}
In the line graph stream, every bond attends to every other bond via global self-attention, giving each bond a full receptive field over all other bonds in a single layer.
At each layer $l$, the bond states $\mathbf{z}^{(l)}$ are updated via pre-norm self-attention followed by a feedforward network, both with residual connections:
\begin{align}
    \hat{\mathbf{z}}^{(l)} &= \mathbf{z}^{(l)} + \mathrm{SelfAttn}\!\left(\mathrm{LN}(\mathbf{z}^{(l)})\right), \label{eq:lg_attn}\\
    \mathbf{z}^{(l+1)} &= \hat{\mathbf{z}}^{(l)} + \mathrm{FFN}\!\left(\mathrm{LN}(\hat{\mathbf{z}}^{(l)})\right), \label{eq:lg_ffn}
\end{align}
where $\mathrm{LN}$ denotes layer normalization and $\mathrm{FFN}$ is a two-layer feedforward network.

Since $M = \binom{N}{2}$ grows quadratically with the number of atoms, the $O(M^2)$ cost of standard softmax self-attention would become prohibitive for larger molecules.
We address this by replacing the softmax kernel with a FAVOR+ random-feature approximation~\citep{performer}, reducing the per-layer attention cost to $O(MR)$ where $R \ll M$ is the number of random features.
A detailed empirical analysis of memory and wall-clock time savings is provided in Appendix~\ref{app:scalability}.

\subsection{Incidence-Constrained Cross-Attention}
\label{sec:cross_attn}

Decoupling atom-level and bond-level reasoning into independent streams is only useful if the two streams can communicate: without synchronization, each stream would reason in isolation, unaware of the other's conclusions.
The cross-attention mechanism synchronizes the two streams at every layer, enabling the atom stream to query the latest bond-level conclusions and the bond stream to query the latest atom-level chemical environments.
Crucially, this communication must respect the molecular topology.
An atom's chemical environment is determined solely by the bonds it participates in, and a bond's character is determined solely by its two endpoint atoms.
We formalize this constraint through the \emph{incidence matrix} $\mathbf{I} \in \{0,1\}^{N \times M}$:  $\mathbf{I}_{i, u_{kl}} = \mathbf{1}[i \in \{k, l\}]$.
% \begin{equation}
%     \mathbf{I}_{i, u_{kl}} = \mathbf{1}[i \in \{k, l\}].
%     \label{eq:incidence}
% \end{equation}
At each layer $l$, bidirectional cross-attention is performed with $\mathbf{I}$ as an attention mask.

\paragraph{Atom update from bonds.}
Each atom aggregates information from its incident bonds.
For atom $i$, let $\mathcal{N}_{\mathbf{I}}(i) = \{u : \mathbf{I}_{i,u} = 1\}$ denote its incident line graph nodes.
The update is:
\begin{equation}
    \mathbf{h}^{(l)}_i \leftarrow \mathbf{h}^{(l)}_i + \mathbf{W}_O^{\mathrm{P\!L}}\sum_{u \in \mathcal{N}_{\mathbf{I}}(i)} \alpha^{\mathrm{P\!L}}_{i,u} \, \mathbf{W}_V^{\mathrm{P\!L}} \,\mathbf{z}^{(l)}_u,
    \label{eq:p_from_l}
\end{equation}
\begin{equation}
    \alpha^{\mathrm{P\!L}}_{i,u} = \frac{\exp\!\left((\mathbf{W}_Q^{\mathrm{P\!L}} \mathbf{h}^{(l)}_i)^\top (\mathbf{W}_K^{\mathrm{P\!L}} \mathbf{z}^{(l)}_u)\, / \sqrt{d_x / n_h}\right)}{\sum_{u' \in \mathcal{N}_{\mathbf{I}}(i)} \exp\!\left((\mathbf{W}_Q^{\mathrm{P\!L}} \mathbf{h}^{(l)}_i)^\top (\mathbf{W}_K^{\mathrm{P\!L}} \mathbf{z}^{(l)}_{u'})\, / \sqrt{d_x / n_h}\right)},
    \label{eq:p_from_l_attn}
\end{equation}
where $\mathbf{W}_Q^{\mathrm{P\!L}}, \mathbf{W}_K^{\mathrm{P\!L}}, \mathbf{W}_V^{\mathrm{P\!L}}, \mathbf{W}_O^{\mathrm{P\!L}}$ are learnable parameters.
Chemically, this allows each atom to gather information about the types and interaction patterns of all bonds it participates in, reflecting how an atom's chemical environment (hybridization, oxidation state, partial charge) is governed by its bonding context.

\paragraph{Bond update from atoms.}
Each bond aggregates information from its two endpoint atoms:
\begin{equation}
    \mathbf{z}^{(l)}_{u_{ij}} \leftarrow \mathbf{z}^{(l)}_{u_{ij}} + \mathbf{W}_O^{\mathrm{L\!P}}\sum_{v \in \{i, j\}} \alpha^{\mathrm{L\!P}}_{u_{ij},v} \, \mathbf{W}_V^{\mathrm{L\!P}} \,\mathbf{h}^{(l)}_v,
    \label{eq:l_from_p}
\end{equation}
where the attention weights $\alpha^{\mathrm{L\!P}}_{u_{ij},v}$ are computed analogously over the two-element set $\{i,j\}$.
This reflects the chemical principle that a bond's properties (order, length, polarity, and reactivity) are determined by the electronegativity, hybridization, and steric environment of its endpoint atoms.

\subsection{Global Fusion and Output}
\label{sec:output}

\paragraph{Global fusion.}
At the end of each layer $l$, the global state integrates information from both streams:
\begin{equation}
    \mathbf{y}^{(l+1)} = \mathrm{LN}\!\left(\mathbf{y}^{(l)} + \mathbf{W}_g \!\left[\mathbf{y}^{(l)}_{\mathrm{P}} \,\|\, \mathbf{y}^{(l)}_{\mathrm{L}} \,\|\, \overline{\mathbf{h}}^{(l)} \,\|\, \overline{\mathbf{z}}^{(l)}\right]\right),
    \label{eq:global_fusion}
\end{equation}
where $\mathbf{y}^{(l)}_{\mathrm{P}}$ and $\mathbf{y}^{(l)}_{\mathrm{L}}$ are the global outputs from the primal and line graph layers, $\overline{\mathbf{h}}^{(l)}$ and $\overline{\mathbf{z}}^{(l)}$ denote mean-pooled node and bond representations, and $\mathbf{W}_g$ is a learnable projection.
This fusion ensures that the global conditioning, which encodes the spectral information, is continuously enriched with structural knowledge from both atom-level and bond-level representations.

\paragraph{FiLM conditioning.}
At the beginning of each layer, the global state modulates both primal and line graph representations via Feature-wise Linear Modulation (FiLM)~\cite{film, mbgen}:
\begin{equation}
    \mathbf{h}^{(l)} \leftarrow \bigl(\mathbf{1} + \boldsymbol{\gamma}_{\mathrm{P}}(\mathbf{y}^{(l)})\bigr) \odot \mathbf{h}^{(l)} + \boldsymbol{\beta}_{\mathrm{P}}(\mathbf{y}^{(l)}), \quad
    \mathbf{z}^{(l)} \leftarrow \bigl(\mathbf{1} + \boldsymbol{\gamma}_{\mathrm{L}}(\mathbf{y}^{(l)})\bigr) \odot \mathbf{z}^{(l)} + \boldsymbol{\beta}_{\mathrm{L}}(\mathbf{y}^{(l)}),
    \label{eq:film}
\end{equation}
where $\boldsymbol{\gamma}(\cdot)$ and $\boldsymbol{\beta}(\cdot)$ are learned linear functions, and $\odot$ denotes element-wise multiplication.
This allows the spectral conditioning to influence the denoising process at every layer and in both streams.

\paragraph{Edge output.}
After the final layer $K$, the line graph node states are decoded back into bond type predictions.
Each $\mathbf{z}^{(K)}_{u_{ij}}$ is mapped to bond type logits via an output network $f_{\mathrm{out}}: \mathbb{R}^{d_e} \to \mathbb{R}^{b+1}$, placed into the symmetric $N \times N$ matrix, and combined with the initial (noisy) edge features via a residual connection:
\begin{equation}
    \hat{\mathbf{E}}_{ij} = \hat{\mathbf{E}}_{ji} = f_{\mathrm{out}}(\mathbf{z}^{(K)}_{u_{ij}}) + \mathbf{E}^{(0)}_{ij}.
    \label{eq:output}
\end{equation}

\section{Experiments}
\label{sec:experiments}

\subsection{Experimental Setup}
\label{sec:experimental_setup}

\paragraph{Datasets.}
We evaluate on two benchmarks adopted by prior work~\cite{diffms, mbgen}.
\textbf{NPLIB1}~\cite{nplib1} is a natural product tandem mass spectrometry dataset derived from GNPS, where training and test sets share structurally similar molecules (Tanimoto $> 0.85$).
\textbf{MassSpecGym}~\cite{massspecgym} is a large-scale benchmark with a strict scaffold-based split (no test molecule has MCES $< 10$ to any training molecule), representing a more challenging out-of-distribution setting.
The two benchmarks differ substantially in scale and chemical diversity, providing a comprehensive evaluation of generalization ability.

\paragraph{Metrics.}
Following prior work~\cite{diffms, mbgen}, 100 candidate molecules are generated per query spectrum. We report three complementary metrics evaluated at top-$K$ for $K \in \{1, 10\}$:
\textbf{Accuracy}, the fraction of samples for which the ground-truth molecule appears among the top-$K$ candidates (matched by InChI);
\textbf{MCES distance} (Maximum Common Edge Subgraph; lower is better), the minimum subgraph-level structural distance between the top-$K$ candidates and the ground truth~\cite{kretschmer2023small};
and \textbf{Tanimoto similarity}, the maximum Tanimoto coefficient (Morgan fingerprints~\cite{morgan1965generation}, radius 2, 2048 bits) between the top-$K$ candidates and the ground truth.

\paragraph{Baselines.}
We compare against Spec2Mol~\cite{spec2mol}, MIST + Neuraldecipher~\cite{le2020neuraldecipher}, MIST + MSNovelist~\cite{msnovelist,zhao2024train}, MADGEN~\cite{madgen}, DiffMS~\cite{diffms}, MS-BART~\cite{msbart}, MSAnchor~\cite{msanchor}, and MBGen~\cite{mbgen}.
MassSpecGym additionally includes SMILES Transformer, SELFIES Transformer, and Random Chemical Generation~\cite{massspecgym}.

\paragraph{Implementation details.}
Following DiffMS~\cite{diffms}, we adopt a two-stage training paradigm.
The denoising network is first pretrained on fingerprint-conditioned molecular generation using a combined corpus drawn from HMDB~\cite{hmdb}, COCONUT~\cite{coconut}, DSSTox~\cite{dsstox}, and MOSES~\cite{moses}.
The spectrum encoder adopts the MIST architecture~\cite{goldman2023annotating} as in DiffMS, and is initialized from the pretrained weights released by DiffMS~\cite{diffms}.
Both encoder and decoder are then fine-tuned end-to-end on spectrum-conditioned generation.
Full hyperparameters are given in Appendix~\ref{app:hyperparams}.

\subsection{Main Results}
\label{sec:main_results}

\begin{table}[t]
\caption{De novo molecular generation from mass spectra on the NPLIB1~\cite{nplib1} and MassSpecGym~\cite{massspecgym} datasets. Best results are in \textbf{bold}. $^\dagger$Results reproduced from DiffMS. $^\ddagger$Results reproduced from MassSpecGym.}
\label{tab:main}
\centering
\footnotesize
\begin{tabular}{l ccc ccc}
\toprule
\multirow{2}{*}{Model} & \multicolumn{3}{c}{Top-1} & \multicolumn{3}{c}{Top-10} \\
\cmidrule(lr){2-4} \cmidrule(lr){5-7}
 & Accuracy $\uparrow$ & MCES $\downarrow$ & Tanimoto $\uparrow$ & Accuracy $\uparrow$ & MCES $\downarrow$ & Tanimoto $\uparrow$ \\
\midrule
\multicolumn{7}{c}{\textbf{NPLIB1}} \\
\midrule
Spec2Mol$^\dagger$ & 0.00\% & 27.82 & 0.12 & 0.00\% & 23.13 & 0.16 \\
MIST + Neuraldecipher$^\dagger$ & 2.32\% & 12.11 & 0.35 & 6.11\% & 9.91 & 0.43 \\
MIST + MSNovelist$^\dagger$ & 5.40\% & 14.52 & 0.34 & 11.04\% & 10.23 & 0.44 \\
MADGEN & 2.10\% & 20.56 & 0.22 & 2.39\% & 12.69 & 0.27 \\
DiffMS & 8.34\% & 11.95 & 0.35 & 15.44\% & 9.23 & 0.47 \\
MS-BART & 7.45\% & 9.66 & 0.44 & 10.99\% & 8.31 & 0.51 \\
MSAnchor & 8.51\% & 11.12 & 0.38 & 16.90\% & 8.95 & 0.49 \\
MBGen & 12.20\% & 7.72 & 0.41 & 22.29\% & 6.71 & 0.50 \\
\textbf{DualLGD} & \textbf{34.37\%} & \textbf{5.44} & \textbf{0.60} & \textbf{44.96\%} & \textbf{4.12} & \textbf{0.68} \\
\midrule
\multicolumn{7}{c}{\textbf{MassSpecGym}} \\
\midrule
SMILES Transformer$^\ddagger$ & 0.00\% & 79.39 & 0.03 & 0.00\% & 52.13 & 0.10 \\
MIST + MSNovelist$^\dagger$ & 0.00\% & 45.55 & 0.06 & 0.00\% & 30.13 & 0.15 \\
SELFIES Transformer$^\ddagger$ & 0.00\% & 38.88 & 0.08 & 0.00\% & 26.87 & 0.13 \\
Spec2Mol$^\dagger$ & 0.00\% & 37.76 & 0.12 & 0.00\% & 29.40 & 0.16 \\
MIST + Neuraldecipher$^\dagger$ & 0.00\% & 33.19 & 0.14 & 0.00\% & 31.89 & 0.16 \\
Random Chemical Generation$^\ddagger$ & 0.00\% & 21.11 & 0.08 & 0.00\% & 18.26 & 0.11 \\
MADGEN & 1.31\% & 27.47 & 0.20 & 1.54\% & 16.84 & 0.26 \\
DiffMS & 2.30\% & 18.45 & 0.28 & 4.25\% & 14.73 & 0.39 \\
MS-BART & 1.07\% & 16.47 & 0.23 & 1.11\% & 15.12 & 0.28 \\
MSAnchor & 2.68\% & 16.57 & 0.32 & 4.67\% & 14.12 & 0.41 \\
MBGen & 7.58\% & 13.25 & 0.38 & 12.54\% & 10.16 & 0.47 \\
\textbf{DualLGD} & \textbf{23.89\%} & \textbf{10.11} & \textbf{0.52} & \textbf{33.08\%} & \textbf{7.69} & \textbf{0.61} \\
\bottomrule
\end{tabular}
\end{table}

Table~\ref{tab:main} compares DualLGD against all baselines on both benchmarks.
DualLGD achieves a top-1 accuracy of \textbf{34.37\%} on NPLIB1 and \textbf{23.89\%} on MassSpecGym, surpassing the previous state of the art (12.20\% and 7.58\%) by approximately \textbf{3$\boldsymbol{\times}$} on both benchmarks.
The improvement is consistent across all metrics: on NPLIB1, MCES decreases from 7.72 to 5.44 (a 29.5\% reduction in structural distance) and Tanimoto improves from 0.41 to 0.60; on MassSpecGym, MCES decreases from 13.25 to 10.11 and Tanimoto improves from 0.38 to 0.52.
At top-10, DualLGD achieves 44.96\% accuracy on NPLIB1 and 33.08\% on MassSpecGym, more than doubling the previous best (22.29\% and 12.54\%).
The scale of improvement is striking given that the best prior method already represented a substantial advance over earlier approaches: on NPLIB1, the gap between the previous two best methods was 3.69 percentage points (8.51\% $\to$ 12.20\%), whereas DualLGD widens this margin to 22.17 percentage points.
This suggests that the dual-stream paradigm, which decouples atom-level and bond-level reasoning into independent representation spaces with per-layer topology-aware synchronization, addresses a key architectural bottleneck in existing single-stream architectures used by all prior methods, enabling qualitatively better bond-level reasoning that translates to dramatically more accurate molecular generation.

To understand how DualLGD's performance varies with structural descriptors, we stratify the NPLIB1 test set along four dimensions: (a)~heavy-atom count, (b)~ring count, (c)~aromatic ring count, and (d)~heteroatom count (Appendix~\ref{app:stratified_analysis}).
To provide further insight into DualLGD's generation quality, we present qualitative results in Appendix~\ref{app:qualitative}.

\subsection{Ablation Studies}
\label{sec:ablation}

\paragraph{Pre-training strategy.}

\begin{table}[t]
\caption{Ablation study on pre-training strategies (NPLIB1). All variants are fine-tuned end-to-end on spectrum-conditioned generation. \textbf{Enc.\ PT}: encoder pre-training; \textbf{Dec.\ PT}: decoder pre-training.}
\label{tab:ablation}
\centering
\footnotesize
\begin{tabular}{cc ccc ccc}
\toprule
\multicolumn{2}{c}{Pre-training Modules} & \multicolumn{3}{c}{Top-1} & \multicolumn{3}{c}{Top-10} \\
\cmidrule(lr){1-2} \cmidrule(lr){3-5} \cmidrule(lr){6-8}
Enc. PT & Dec. PT & Accuracy $\uparrow$ & MCES $\downarrow$ & Tanimoto $\uparrow$ & Accuracy $\uparrow$ & MCES $\downarrow$ & Tanimoto $\uparrow$ \\
\midrule
- & - & 25.53\% & 7.21 & 0.50 & 33.75\% & 5.70 & 0.58 \\
\checkmark & - & 28.89\% & 6.69 & 0.53 & 37.26\% & 5.28 & 0.60 \\
- & \checkmark & 33.13\% & 5.91 & 0.57 & \textbf{45.58}\% & 4.23 & 0.67 \\
\checkmark & \checkmark & \textbf{34.37\%} & \textbf{5.44} & \textbf{0.60} & 44.96\% & \textbf{4.12} & \textbf{0.68} \\
\bottomrule
\end{tabular}
\end{table}

Table~\ref{tab:ablation} isolates the contributions of encoder and decoder pre-training on NPLIB1.
Decoder pre-training provides the largest individual gain, raising top-1 accuracy from 25.53\% to 33.13\% (+7.60 percentage points) and reducing MCES from 7.21 to 5.91.
This is expected: fingerprint-conditioned pre-training directly teaches the denoising network to reconstruct molecular graphs from conditioning vectors, providing a strong initialization for the spectrum-conditioned task.
Encoder pre-training alone yields a meaningful improvement (+3.36 pp), and when combined with decoder pre-training further improves accuracy to 34.37\% and MCES to 5.44, suggesting that a better-aligned spectrum encoder complements the pretrained decoder.
Critically, \textbf{DualLGD without any pre-training already achieves 25.53\% top-1 accuracy, more than doubling the previous best fully pretrained model (12.20\%)} (cf.\ Table~\ref{tab:main}).
This demonstrates that the performance leap originates primarily from the DualLGD architecture itself, not the training recipe.
Pre-training amplifies this architectural advantage but is not its source.
As a further reference, when trained from scratch without any pre-training, DiffMS achieves only 0.50\% and MBGen achieves 0.00\% top-1 accuracy, compared to DualLGD's 25.53\%.
The gap exceeds an order of magnitude, underscoring that the dual-stream architecture fundamentally changes the model's ability to learn effective bond-level representations, independent of the training recipe.

\paragraph{Dual-stream architecture and cross-attention.}

\begin{table}[t]
\caption{Ablation study on the dual-stream architecture and cross-attention (NPLIB1). All models are trained from scratch without pre-training. \textbf{A$\to$B}: atom representations update bond representations only; \textbf{B$\to$A}: bond representations update atom representations only.}
\label{tab:ablation_crossattn}
\centering
\footnotesize
\begin{tabular}{l ccc ccc}
\toprule
\multirow{2}{*}{Variant} & \multicolumn{3}{c}{Top-1} & \multicolumn{3}{c}{Top-10} \\
\cmidrule(lr){2-4} \cmidrule(lr){5-7}
& Accuracy $\uparrow$ & MCES $\downarrow$ & Tanimoto $\uparrow$ & Accuracy $\uparrow$ & MCES $\downarrow$ & Tanimoto $\uparrow$ \\
\midrule
w/o Line Graph Stream & 2.24\% & 14.71 & 0.24 & 6.23\% & 10.72 & 0.35 \\
w/o Cross-Attention & 0.62\% & 20.18 & 0.15 & 1.37\% & 16.52 & 0.22 \\
\midrule
A$\to$B only & 2.37\% & 19.07 & 0.20 & 3.99\% & 13.89 & 0.29 \\
B$\to$A only & 0.00\% & 27.39 & 0.13 & 1.00\% & 23.01 & 0.19 \\
w/o Incidence Mask & 0.25\% & 20.74 & 0.14 & 1.00\% & 16.88 & 0.20 \\
\midrule
\textbf{DualLGD (Full)} & \textbf{25.53\%} & \textbf{7.21} & \textbf{0.50} & \textbf{33.75\%} & \textbf{5.70} & \textbf{0.58} \\
\bottomrule
\end{tabular}
\end{table}

Table~\ref{tab:ablation_crossattn} ablates the core architectural components, with all variants trained from scratch.
Strikingly, operating the line graph stream without cross-attention (w/o Cross-Attention, 0.62\%) performs \emph{worse} than removing the line graph stream entirely (2.24\%): an unsynchronized bond stream not only fails to help but actively interferes with atom-level learning.
This confirms that a dedicated bond representation space creates value only when coupled with a synchronization mechanism; the two components are mutually enabling rather than independently additive.
Among the directional ablations, B$\to$A only achieves 0.00\% accuracy, because bond representations that have never been grounded in atom-level context carry no useful signal back to atoms; A$\to$B only fares better (2.37\%) but leaves the loop open, yielding performance comparable to the no-bond-stream baseline.
Full bidirectional cross-attention closes this loop, enabling both streams to co-refine their representations at every layer and yielding a 10$\times$ improvement over the best unidirectional variant.
Removing the incidence mask collapses performance to near zero, demonstrating that the topology constraint is not a computational convenience but an essential structural prior encoding the chemical principle that an atom's environment is determined solely by its incident bonds.
Together, these results confirm that DualLGD's architectural gains arise from all three design choices working in concert: a dedicated bond representation space, full bidirectional cross-attention, and the incidence mask as a topology-aware structural prior.

\subsection{Efficient Long-Range Reverse Sampling}
\label{app:elrt}

\begin{figure}[t]
    \centering
    \includegraphics[width=\textwidth]{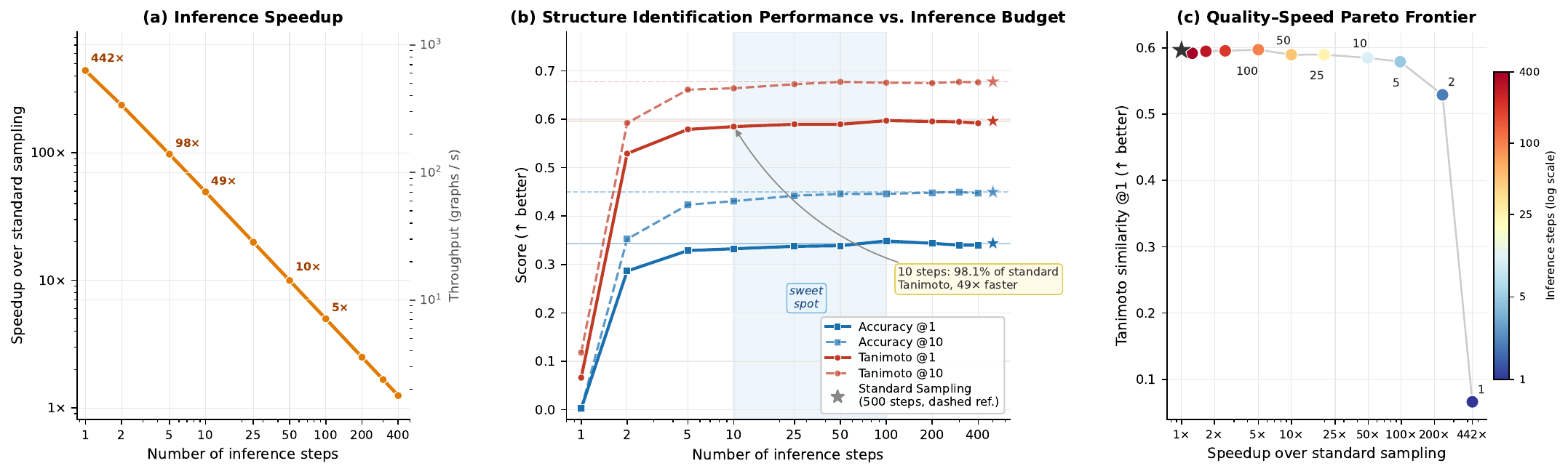}
    \caption{
    % Speed-quality trade-off of efficient long-range reverse sampling on NPLIB1.
    % \textbf{(a)} Wall-clock speedup relative to standard 500-step sampling as a function of the number of inference steps; annotations mark selected speedup values.
    % \textbf{(b)} Top-1/10 accuracy and Tanimoto similarity as a function of inference steps; stars mark the standard 500-step baseline and dashed horizontal lines indicate the corresponding reference values. The shaded region (10--100 steps) identifies the practical sweet spot.
    % \textbf{(c)} Quality-speed Pareto frontier: Tanimoto@1 vs.\ wall-clock speedup; each point is colored by inference step count (log scale). No retraining is required.
    Speed–quality trade-off of efficient long-range reverse sampling on NPLIB1.
    \textbf{(a)} Wall-clock speedup over standard 500-step sampling versus the number of inference steps.
    \textbf{(b)} Top-1/10 accuracy and Tanimoto similarity versus inference steps, with the 500-step baseline shown for reference.
    \textbf{(c)} Quality–speed Pareto frontier, plotting Tanimoto@1 against wall-clock speedup.
    }
    \label{fig:elrt}
\end{figure}

Standard discrete diffusion inference requires $T$ denoising network evaluations, one per timestep.
We reduce this cost without any retraining by combining two properties of the DiGress marginal categorical noise process.

% \paragraph{Closure of marginal transition matrices.}
% The family of marginal categorical transition matrices used in D3PM~\citep{austin2021structured} and DiGress~\citep{digress} is closed under composition: the forward transition $q(G_t \mid G_s)$ for any stride $t - s \geq 1$ can be computed in closed form from the noise schedule alone.
% The reverse posterior $q(G_s \mid G_t, G_0)$ then follows from Bayes' rule using three such closed-form matrices.
% Consequently, $p_\theta(G_s \mid G_t)$ can be evaluated exactly for any $s < t$, with no approximation beyond the model's own prediction $\hat{G}_0 = \phi_\theta(G_t, t, \mathbf{y})$.

% \paragraph{Long-range jump sampling.}
% Building on this closure, DNDM~\citep{chen2024fast} proposes replacing the standard unit-stride reverse chain with a non-Markovian trajectory that jumps directly between a selected subset of $T' \ll T$ timesteps.
% We adopt this scheme: given a monotonically decreasing subsequence $T = \tau_1 > \tau_2 > \cdots > \tau_{T'} = 0$, each reverse step evaluates the denoising network once at $\tau_i$ and computes $p_\theta(G_{\tau_{i+1}} \mid G_{\tau_i})$ exactly via the closed-form $(\tau_i - \tau_{i+1})$-step posterior.
% The training objective is unchanged.
% The only trade-off is that larger jumps place a heavier denoising burden on each network evaluation, potentially amplifying prediction errors; we quantify this in Figure~\ref{fig:elrt}.

\textbf{Closure of marginal transition matrices.}
The marginal categorical transition matrices used in D3PM~\citep{austin2021structured} and DiGress~\citep{digress} are closed under composition.
In our setting, atom types $\mathbf{X}$ are fixed by the molecular formula and diffusion is applied only to bond variables $\mathbf{E}$, i.e., $G_t=(\mathbf{X},\mathbf{E}_t)$.
Thus, for any $0\le s<t\le T$, the multi-step transition $q(\mathbf{E}_t\mid \mathbf{E}_s)$ is available in closed form.
Under the factorized categorical corruption process, the skipped posterior factorizes over unordered atom pairs:
\begin{equation}
q(\mathbf{E}_{s,ij}\mid \mathbf{E}_{t,ij},\mathbf{E}_{0,ij})
=
\frac{
q(\mathbf{E}_{t,ij}\mid \mathbf{E}_{s,ij})q(\mathbf{E}_{s,ij}\mid \mathbf{E}_{0,ij})
}{
q(\mathbf{E}_{t,ij}\mid \mathbf{E}_{0,ij})
},
\end{equation}
with normalization over $\mathbf{E}_{s,ij}$.
At inference time, the network predicts $\hat p_\theta(\mathbf{E}_0\mid \mathbf{E}_t,\mathbf{X},\mathbf y)$.
For each pair $i<j$, we marginalize over all clean bond types $\tilde e_0\in\{0,1,\ldots,b\}$ to obtain
\begin{equation}
p_\theta(\mathbf{E}_{s,ij}\mid \mathbf{E}_t,\mathbf{X},\mathbf y)
=
\sum_{\tilde e_0\in\{0,1,\ldots,b\}}
q(\mathbf{E}_{s,ij}\mid \mathbf{E}_{t,ij},\tilde e_0)
\hat p_\theta(\mathbf{E}_{0,ij}=\tilde e_0\mid \mathbf{E}_t,\mathbf{X},\mathbf y).
\end{equation}
The skipped posterior is exact under the forward process; the approximation comes only from the learned clean-bond prediction.

\textbf{Long-range jump sampling.}
Inspired by DNDM~\citep{chen2024fast}, we use a training-free skipped sampler for the DiGress marginal categorical process.
Instead of assigning variable-specific transition times, we choose a global decreasing subsequence
$T=\tau_0>\tau_1>\cdots>\tau_J=0$, where $J$ is the number of denoising network evaluations.
For each $j=0,\ldots,J-1$, we evaluate the network at $\tau_j$, construct
$p_\theta(\mathbf{E}_{\tau_{j+1}}\mid \mathbf{E}_{\tau_j},\mathbf{X},\mathbf y)$
using the skipped posterior above, and sample $\mathbf{E}_{\tau_{j+1}}$.
This generalizes unit-step sampling from $s=t-1$ to arbitrary $s<t$ without retraining.
Larger jumps reduce inference cost but increase the denoising burden per evaluation; we quantify this trade-off in Figure~\ref{fig:elrt}.

% \paragraph{Results.}
Figure~\ref{fig:elrt} reports the speed-quality trade-off on NPLIB1.
With only 10 inference steps, DualLGD achieves a $49\times$ wall-clock speedup while retaining 98.1\% of the standard 500-step Tanimoto similarity.
Performance degrades sharply only below 2 steps and saturates above roughly 50 steps, identifying 10--100 steps as a practical sweet spot that delivers substantial inference acceleration at negligible quality cost.
For applications where maximum throughput is required, 2 steps still retains 88.8\% of the standard Tanimoto similarity while achieving a $237\times$ wall-clock speedup.
% These results demonstrate that DualLGD's discrete diffusion process is well-suited to post-hoc inference acceleration via long-range jump sampling.

\subsection{What Does Cross-Attention Learn?}
\label{sec:cross_attn_interp}

\begin{figure}[t]
    \centering
    \includegraphics[width=0.9\textwidth]{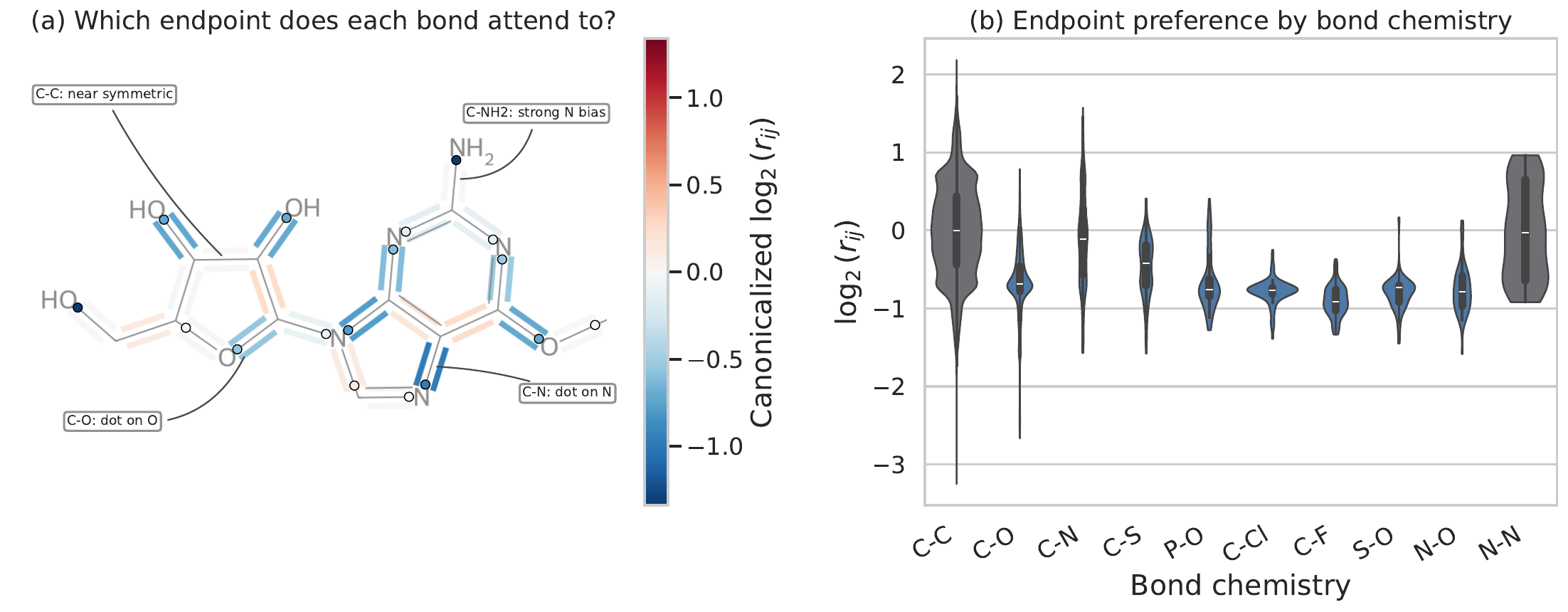}
    % \caption{Cross-attention learns electronegativity-aware endpoint preference without explicit supervision.
    % (a) For each bond in a representative molecule (21 heavy atoms), the color encodes $\log_2(r_{ij})$, where $r_{ij}$ is the asymmetry ratio.
    % Dots mark the more-attended endpoint.
    % Heterogeneous bonds (C--O, C--N) preferentially attend to the more electronegative atom, while C--C bonds remain near symmetric.
    % (b) Test-set-wide statistics grouped by bond chemistry.
    % For heterogeneous bonds, $r_{ij}$ is the ratio of attention to the less electronegative endpoint over attention to the more electronegative endpoint; for homonuclear bonds (C--C, N--N), orientation is randomized.
    % % All heterogeneous bond types show a systematic negative bias in $\log_2 r_{ij}$, with magnitude increasing with the electronegativity difference between the two endpoint atoms.}
    % All heterogeneous bond types show a systematic negative bias in $\log_2 r_{ij}$, indicating preferential attention to the more electronegative endpoint, with bond-type-dependent magnitude.}
    \caption{Cross-attention learns electronegativity-aware endpoint preference without explicit supervision.
    (a) In a representative molecule with 21 heavy atoms, color encodes the log attention asymmetry ratio, and dots mark the more-attended endpoint. Heterogeneous bonds such as C--O and C--N tend to attend to the more electronegative atom, while C--C bonds remain nearly symmetric.
    (b) Test-set-wide statistics grouped by bond chemistry; negative values indicate stronger attention to the more electronegative endpoint, while homonuclear bonds are centered near zero.}
    \label{fig:bond_asymmetry}
\end{figure}

% Table~\ref{tab:ablation_crossattn} establishes incidence-constrained bidirectional cross-attention as indispensable for DualLGD's performance.
% A natural question follows: what information does cross-attention transfer between atoms and bonds?
We analyze the atom$\to$bond direction of cross-attention (Eq.~\eqref{eq:l_from_p}), in which each bond distributes attention weights over its two endpoint atoms.
For a bond $(i,j)$, let $\alpha^{\mathrm{LP}}_{u_{ij},v}$ denote the attention weight assigned to endpoint $v \in \{i, j\}$, and let $i_{\mathrm{high}}$, $i_{\mathrm{low}}$ denote the more and less electronegative endpoint, respectively.
We define the \emph{asymmetry ratio}
   $ r_{ij} = \frac{\alpha^{\mathrm{LP}}_{u_{ij},\, i_{\mathrm{low}}}}{\alpha^{\mathrm{LP}}_{u_{ij},\, i_{\mathrm{high}}}},$
so that $\log_2 r_{ij} < 0$ indicates a preference for the more electronegative endpoint.
For heterogeneous bonds (e.g., C--O, C--N, C--F, C--Cl), we orient the ratio toward the more electronegative endpoint; for homonuclear bonds (C--C, N--N), orientation is randomized.

Figure~\ref{fig:bond_asymmetry} summarizes the results across the NPLIB1 test set.
Two patterns emerge.
First, all heterogeneous bond types exhibit a systematic negative bias in $\log_2 r_{ij}$: bonds consistently allocate more attention to the more electronegative endpoint, while homonuclear bonds are centered near zero, as expected under random orientation.
% Second, the magnitude of the bias increases monotonically with the electronegativity difference between the two endpoints: C--F and C--Cl show the largest shifts, while C--N, which has the smallest electronegativity contrast among heterogeneous types, shows the weakest but still significant bias.
Second, the bias generally favors the more electronegative endpoint, but its magnitude varies across bond types and is not strictly monotonic with electronegativity difference. C--F and C--Cl show large shifts, while other heterogeneous bonds such as C--O, C--N, and C--S exhibit smaller but still systematic biases. These differences suggest that attention routing reflects electronegativity together with bond-specific chemical and topological context.
Critically, no supervision relating to electronegativity was provided during training.
The more electronegative atom exerts a larger influence on a bond's polarity, reactivity, and cleavage tendency; attending preferentially to that endpoint therefore provides a richer chemical context for updating the bond representation.
The cross-attention mechanism spontaneously learns this electronegativity-aware information routing from molecular graph structure alone.

% This finding indicates that cross-attention not only proves indispensable in ablation experiments, but also learns an information-routing strategy aligned with fundamental chemical principles, providing an interpretable basis for DualLGD's denoising accuracy.

\section{Conclusion}
\label{sec:conclusion}

We have presented DualLGD, a dual-stream diffusion framework that reformulates molecular graph denoising as the alternating solution of atom-level and bond-level subproblems.
% The core insight is that the bottleneck of existing methods lies not in the strength of their edge aggregation kernels, but in the single-stream architectural paradigm where atom--bond information synchronization can only occur implicitly across layers.
DualLGD promotes bonds to first-class nodes in a line graph and synchronizes the two streams at every layer through incidence-constrained bidirectional cross-attention that respects molecular topology.
On NPLIB1 and MassSpecGym, DualLGD achieves approximately $3\times$ improvement in top-1 accuracy over the previous state of the art, with ablation studies confirming that the dual-stream architecture, rather than the training recipe, is the primary source of this gain.
More broadly, DualLGD suggests that topology-synchronized atom and bond streams are an effective design principle for molecular generation. This design improves bond-level reasoning but adds computational cost for large or complex molecules, motivating future work on more efficient bond-space modeling, adaptive synchronization, and stereochemical or 3D extensions.

\begin{ack}
This work was supported in part by U.S. National Science Foundation (2348391, PI Xu) and UNC Charlotte internal funding (Du).
\end{ack}

\small
\bibliographystyle{plainnat}
\bibliography{references}

%%%%%%%%%%%%%%%%%%%%%%%%%%%%%%%%%%%%%%%%%%%%%%%%%%%%%%%%%%%%

\appendix

\section{Hyperparameters}
\label{app:hyperparams}

Table~\ref{tab:arch-hyper} summarizes the architectural hyperparameters of the DualLGD denoising network.
Table~\ref{tab:train-hyper} reports the training configurations for each stage.

\begin{table}[h]
\caption{Architectural hyperparameters of the DualLGD denoising network.}
\label{tab:arch-hyper}
\centering
\small
\begin{tabular}{@{}ll@{}}
\toprule
Hyperparameter & Value \\
\midrule
\multicolumn{2}{@{}l}{\textit{Dual-stream transformer}} \\[2pt]
\quad Number of layers $K$ & 5 \\
\quad Node hidden dim $d_x$ & 256 \\
\quad Edge hidden dim $d_e$ & 64 \\
\quad Global hidden dim $d_y$ & 1024 \\
\quad Attention heads (primal / dual / cross) & 8 / 8 / 8 \\
\quad FFN hidden dims (node / edge / global) & 256 / 128 / 2048 \\
\quad Dropout / attention dropout / drop path & 0.1 / 0.1 / 0.1 \\[4pt]
\multicolumn{2}{@{}l}{\textit{Diffusion process}} \\[2pt]
\quad Number of diffusion steps $T$ & 500 \\
\quad Noise schedule & Cosine \\
\quad Transition type & Marginal \\
\bottomrule
\end{tabular}
\end{table}

\begin{table}[h]
\caption{Training hyperparameters for each stage.}
\label{tab:train-hyper}
\centering
\small
\begin{tabular}{@{}lccc@{}}
\toprule
& \multirow{2}{*}{Decoder pretraining} & \multicolumn{2}{c}{End-to-end fine-tuning} \\
\cmidrule(lr){3-4}
& & NPLIB1 & MassSpecGym \\
\midrule
Batch size & 16 & 16 & 16 \\
Learning rate & $1.5 \times 10^{-3}$ & $2 \times 10^{-4}$ & $2 \times 10^{-4}$ \\
Epochs & 75 & 50 & 15 \\
Optimizer & \multicolumn{3}{c}{AdamW} \\
Scheduler & \multicolumn{3}{c}{OneCycleLR ($\texttt{pct\_start} = 0.3$)} \\
Weight decay & $10^{-12}$ & $10^{-12}$ & $10^{-12}$ \\
Gradient clipping & --- & 1.0 & 1.0 \\
\bottomrule
\end{tabular}
\end{table}

\paragraph{Training compute.}
Decoder pre-training takes approximately 140 minutes per epoch on 4 NVIDIA H200 GPUs.
End-to-end fine-tuning on a single H200 takes approximately 46 seconds per epoch on NPLIB1 and 22 minutes per epoch on MassSpecGym.

\section{Parameter Comparison}
\label{app:params}

\begin{table}[h]
\caption{Parameter counts of the denoising networks. All models share the same spectrum encoder and diffusion framework; only the denoising network architecture differs.}
\label{tab:params}
\centering
\begin{tabular}{l r r}
\toprule
Model & Parameters (M) & Relative to DiffMS \\
\midrule
DiffMS & 53.18 & --- \\
MBGen & 62.38 & +17.3\% \\
DualLGD & 67.56 & +27.0\% \\
\bottomrule
\end{tabular}
\end{table}

Table~\ref{tab:params} compares the parameter counts of the three denoising networks.
DualLGD adds 14.38M parameters (+27.0\%) over DiffMS.
The dominant source is the global fusion module (12.13M), which aggregates information from both streams at each layer.
The core architectural innovations, the line graph stream (0.58M) and the bidirectional cross-attention modules (1.03M), together account for less than 1.6M parameters, under 2.4\% of the total model.
MBGen adds 9.20M parameters (+17.3\%) over DiffMS through its many-body attention layers.

Despite this modest difference in model size, DualLGD achieves approximately $3\times$ improvement in top-1 accuracy over the previous state of the art on both benchmarks (cf.\ Table~\ref{tab:main}), indicating that the performance gain stems from the architectural design rather than additional model capacity.

\section{Scalability Analysis}
\label{app:scalability}

\begin{figure}[h]
    \centering
    \includegraphics[width=\textwidth]{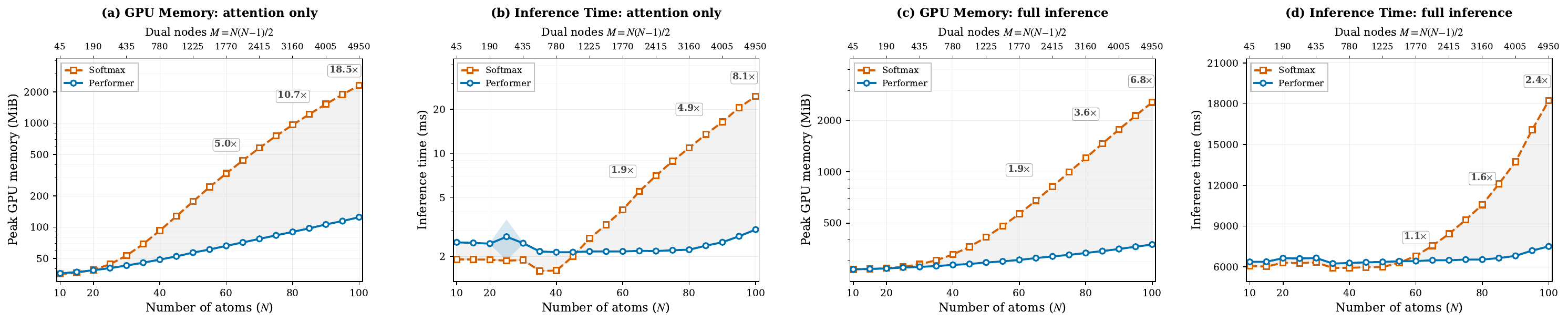}
    \caption{Scalability of the line graph stream self-attention on a single NVIDIA H200 GPU.
    \textbf{(a--b)}~Isolated line graph attention layers: peak GPU memory and per-step inference time as a function of atom count $N$.
    \textbf{(c--d)}~Full 500-step denoising inference loop: end-to-end memory and wall-clock time.
    The top axis shows the number of line graph nodes $M = N(N{-}1)/2$; shaded regions highlight the gap between implementations. Ratios are annotated at $N = 60$, $80$, and $100$.}
    \label{fig:scalability}
\end{figure}

Since the number of line graph nodes $M = \binom{N}{2}$ grows quadratically with the number of atoms, the choice of self-attention implementation in the line graph stream has significant implications for scalability.
We compare two model variants that differ only in this implementation: standard softmax attention ($\mathcal{O}(M^2)$) and an efficient linear approximation via FAVOR+~\cite{performer} ($\mathcal{O}(MR)$, $R = 128$ orthogonal random features), benchmarked on a single NVIDIA H200 GPU across $N = 10$--$100$ atoms ($M = 45$--$4{,}950$ line graph nodes).

\paragraph{Attention-only comparison (Figure~\ref{fig:scalability}a--b).}
Benchmarking the line graph attention layers in isolation, the softmax variant exhibits the expected $\mathcal{O}(M^2)$ scaling: at $N = 100$ ($M = 4{,}950$), peak memory reaches 2.3\,GiB and per-step inference time grows to 24.5\,ms.
The linear approximation scales near-linearly, consuming only 125\,MiB of memory and 3.0\,ms per step at the same size, yielding a \textbf{18.5$\boldsymbol{\times}$} memory reduction and \textbf{8.1$\boldsymbol{\times}$} speedup.
This gap widens steadily with molecular size: at $N = 60$ the ratios are already $5.0\times$ and $1.9\times$, growing to $10.7\times$ and $4.9\times$ at $N = 80$.

\paragraph{Full inference loop (Figure~\ref{fig:scalability}c--d).}
Under realistic deployment conditions (a 500-step denoising loop with edge resampling at each step), the softmax variant at $N = 100$ requires 2.6\,GiB of GPU memory and 18.2\,s of wall-clock time, while the linear approximation completes inference in 7.5\,s using only 375\,MiB (\textbf{6.8$\boldsymbol{\times}$} memory reduction, \textbf{2.4$\boldsymbol{\times}$} end-to-end speedup).
The time speedup is smaller than in the attention-only case (2.4$\times$ vs.\ 8.1$\times$) because the full model includes components shared by both variants that are independent of the attention implementation: the primal stream ($\mathcal{O}(N^2)$), the incidence-constrained cross-attention, FiLM conditioning, global fusion, and per-step edge resampling.
These components constitute a fixed time budget that the linear approximation does not reduce.
As $N$ grows and the $\mathcal{O}(M^2)$ attention cost becomes dominant, the end-to-end speedup increases steadily (from $1.1\times$ at $N = 60$ to $1.6\times$ at $N = 80$ and $2.4\times$ at $N = 100$), and will continue to grow for larger molecules.
The memory ratio (6.8$\times$) more closely tracks the attention-only ratio because peak GPU memory is primarily consumed by the $M \times M$ attention score matrix that softmax attention must materialize; the linear approximation replaces this with an $M \times R$ matrix ($R \ll M$), and the shared components contribute comparatively little to peak memory.

\paragraph{Accuracy comparison.}

\begin{table}[h]
\caption{Generation accuracy of the two line graph attention implementations on NPLIB1, both trained from scratch without pre-training under otherwise identical settings.}
\label{tab:softmax_vs_favor}
\centering
\footnotesize
\begin{tabular}{l ccc ccc}
\toprule
\multirow{2}{*}{Line graph attention} & \multicolumn{3}{c}{Top-1} & \multicolumn{3}{c}{Top-10} \\
\cmidrule(lr){2-4} \cmidrule(lr){5-7}
& Accuracy $\uparrow$ & MCES $\downarrow$ & Tanimoto $\uparrow$ & Accuracy $\uparrow$ & MCES $\downarrow$ & Tanimoto $\uparrow$ \\
\midrule
Softmax (exact, $\mathcal{O}(M^2)$) & 26.03\% & 7.12 & 0.50 & 35.37\% & 5.58 & 0.59 \\
FAVOR+ (linear, $\mathcal{O}(MR)$) & 25.53\% & 7.21 & 0.50 & 33.75\% & 5.70 & 0.58 \\
\bottomrule
\end{tabular}
\end{table}

The scalability gains above are only meaningful if the linear approximation does not compromise generation quality.
We therefore train both variants from scratch under identical settings and report their accuracy in Table~\ref{tab:softmax_vs_favor}.
The FAVOR+ approximation matches exact softmax attention to within 0.50 percentage points in top-1 accuracy (25.53\% vs.\ 26.03\%) and 1.62 percentage points in top-10 accuracy (33.75\% vs.\ 35.37\%), with essentially identical Tanimoto similarity and a marginal MCES gap ($\Delta \leq 0.12$).
This negligible accuracy cost is far outweighed by the order-of-magnitude memory and runtime savings reported above, justifying FAVOR+ as the default attention kernel for the line graph stream.

These results confirm that an efficient attention implementation is necessary for the practical viability of the line graph stream at typical molecular sizes.
At $N = 100$ ($M \approx 4{,}950$), standard softmax attention would be prohibitively expensive in both memory and time.

%%%%%%%%%%%%%%%%%%%%%%%%%%%%%%%%%%%%%%%%%%%%%%%%%%%%%%%%%%%%

\section{Stratified Analysis by Structural Descriptors}
\label{app:stratified_analysis}

\begin{figure}[t]
    \centering
    \includegraphics[width=\textwidth]{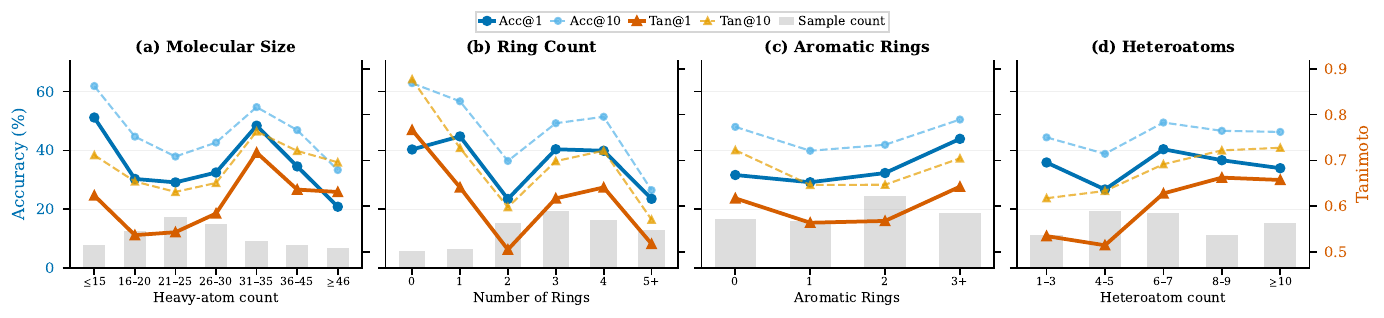}
    \caption{Performance of DualLGD on NPLIB1 stratified by structural descriptors: (a)~heavy-atom count, (b)~ring count, (c)~aromatic ring count, and (d)~heteroatom count. Left axes show accuracy (\%), right axes show Tanimoto similarity. Gray bars indicate the number of samples in each bin.}
    \label{fig:performance_by_property}
\end{figure}

To understand how DualLGD's performance varies with structural descriptors, we stratify the NPLIB1 test set along four dimensions (Figure~\ref{fig:performance_by_property}).

\paragraph{Molecular size (a).}
Small molecules ($\leq$15 heavy atoms) are the easiest, with top-1 accuracy exceeding 50\%.
Performance decreases for medium-sized molecules (16--30 atoms), yet remains well above 25\% across all bins, still far surpassing the previous state-of-the-art overall accuracy of 12.20\%.
Notably, the 31--35 atom bin exhibits a secondary peak (top-1 accuracy ${\sim}$48\%), likely reflecting the prevalence of structurally regular terpenoid and steroidal scaffolds in natural products, whose rigid ring systems and constrained bond angles are well suited to the line graph representation.
For the largest molecules ($\geq$46 atoms), accuracy decreases gracefully to ${\sim}$20\%, while Tanimoto similarity remains above 0.6, indicating that even when the exact structure is not recovered, the generated molecules preserve the core scaffold.

\paragraph{Ring count (b).}
Acyclic molecules achieve the highest top-10 accuracy (${\sim}$63\%) and Tanimoto (${\sim}$0.88), consistent with their smaller combinatorial space of valid bond configurations.
Molecules with 3--4 rings, which constitute nearly half of the NPLIB1 test set, maintain top-1 accuracy around 40\%, demonstrating strong performance on the most representative natural product scaffolds.
Accuracy decreases for molecules with 5 or more rings (${\sim}$24\%), reflecting the inherent difficulty of resolving complex fused ring systems.

\paragraph{Aromatic rings (c).}
A notable trend emerges: accuracy \emph{increases} with the number of aromatic rings, rising from ${\sim}$32\% (no aromatic rings) to ${\sim}$44\% (3 or more).
This aligns directly with the design motivation of the line graph stream: aromatic systems involve extended patterns of alternating single and double bonds, and the direct bond-bond interactions captured by the line graph enable effective learning of $\pi$-electron delocalization and conjugation patterns.

\paragraph{Heteroatoms (d).}
Performance is relatively stable across heteroatom counts, with a peak around 6--7 heteroatoms (top-1 accuracy ${\sim}$40\%), corresponding to the oxygen- and nitrogen-rich scaffolds common in natural products.
Molecules with 10 or more heteroatoms show lower accuracy but maintain Tanimoto similarity above 0.65, confirming that DualLGD captures the correct heteroatom distribution even when exact reconstruction is not achieved.

\paragraph{Summary.}
Across all four dimensions, Tanimoto similarity remains robust (0.55--0.72 at top-1), indicating that DualLGD consistently generates structurally faithful molecules regardless of the property regime.
The positive correlation between aromatic complexity and accuracy provides empirical support for the effectiveness of the line graph stream in capturing bond-level interaction patterns.

\section{Qualitative Analysis}
\label{app:qualitative}

\begin{figure}[t]
    \centering
    \includegraphics[width=\textwidth]{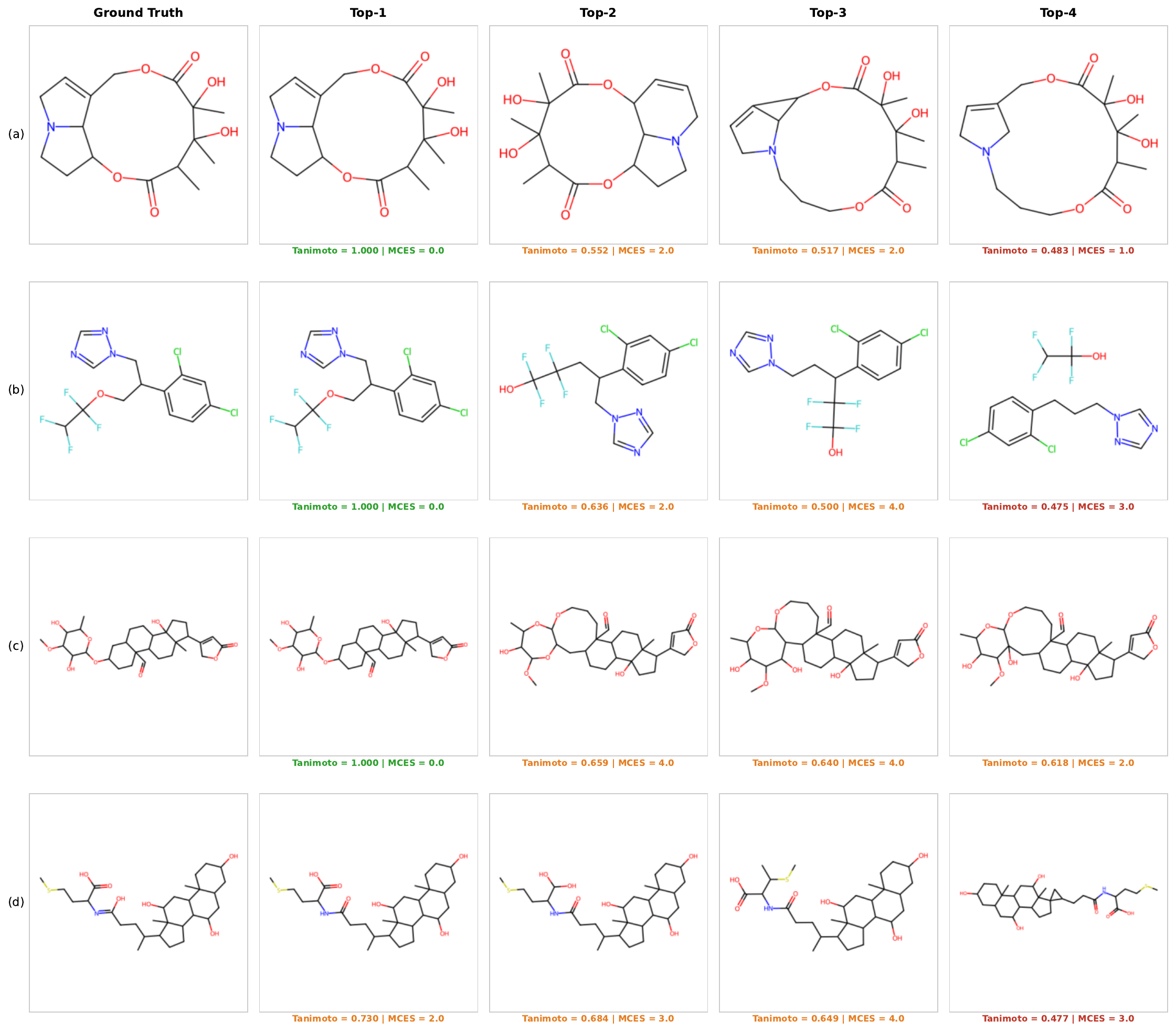}
    \caption{Representative generation results of DualLGD on NPLIB1. Each row shows the ground-truth molecule (left) alongside the top-4 candidates ranked by model likelihood. Tanimoto similarity and MCES distance to the ground truth are reported below each candidate. In examples (a)--(c), the top-1 candidate exactly matches the ground truth (Tanimoto = 1.0, MCES = 0.0). Example (d) illustrates a case where the exact structure is not recovered, yet the top-1 candidate achieves high similarity (Tanimoto = 0.730, MCES = 2.0).}
    \label{fig:molecule_examples}
\end{figure}

\begin{figure}[t]
    \centering
    \includegraphics[width=\textwidth]{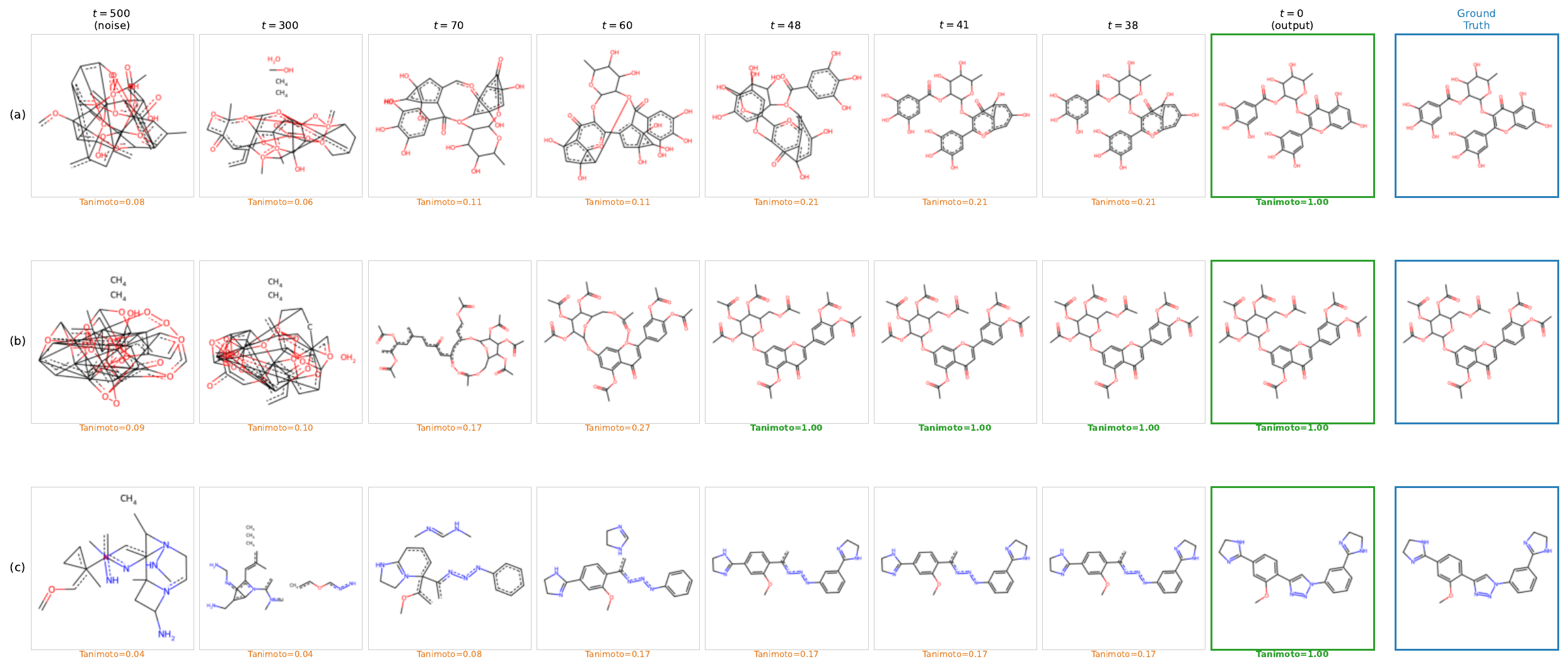}
    \caption{Visualization of the reverse diffusion process for three molecules from NPLIB1. From left to right, the molecular graph evolves from the fully noised state ($t = 500$) to the final output ($t = 0$), with the ground truth shown on the far right. The Tanimoto similarity to the ground truth is annotated at each timestep. During the early stages ($t = 500 \to 70$), the model rapidly eliminates spurious bonds, establishing the overall molecular scaffold. In the later stages ($t < 70$), the model refines local bond types and resolves remaining structural ambiguities to converge on the final prediction.}
    \label{fig:molecule_diffusion}
\end{figure}

To provide further insight into DualLGD's generation quality, we present qualitative results in Figures~\ref{fig:molecule_examples} and~\ref{fig:molecule_diffusion}.

\paragraph{Generation examples.}
Figure~\ref{fig:molecule_examples} shows four representative examples from NPLIB1 with their top-4 candidates.
In examples (a)--(c), DualLGD recovers the exact ground-truth molecular graph as the top-1 prediction, despite the structural complexity of these molecules: (a) is a macrocyclic lactone with a nitrogen-containing ring, (b) contains halogenated aromatic groups with a trifluoromethoxy substituent, and (c) is a large glycosylated natural product with multiple fused rings and extensive heteroatom substitution.
These cases demonstrate DualLGD's ability to correctly resolve diverse chemical motifs, including ring systems, heteroatom substitution patterns, and complex bond-type assignments.
Even when the exact structure is not recovered, as in example (d), the top-ranked candidates remain structurally close to the ground truth (Tanimoto $\geq 0.649$ for the top-3), indicating that the model captures the core molecular scaffold and errs only in local substituent placement.
Moreover, the candidate set exhibits meaningful structural diversity while maintaining high similarity, suggesting that the diffusion process explores chemically plausible alternatives rather than collapsing to trivial perturbations.

\paragraph{Reverse diffusion trajectory.}
Figure~\ref{fig:molecule_diffusion} visualizes the denoising process at selected timesteps for three molecules.
A consistent two-phase pattern emerges across all examples.
In the early phase ($t = 500 \to 70$), the heavily corrupted graph, in which nearly all atom pairs carry noisy bond assignments, is rapidly pruned to establish the global molecular skeleton.
This phase removes the majority of spurious bonds, reducing the dense noise pattern to a recognizable scaffold and reflecting the model's ability to identify which atom pairs should be bonded based on spectral conditioning.
In the refinement phase ($t < 70$), the model resolves finer structural details: distinguishing single from double bonds, assigning aromatic ring patterns, and correcting local connectivity.
Tanimoto similarity correspondingly rises sharply in this stage: examples (a) and (c) reach 1.00 at the final output, while example (b) reaches 1.00 earlier at t = 48 and remains exact thereafter.
This coarse-to-fine behavior is consistent with the diffusion noise schedule, which introduces increasingly subtle perturbations at lower noise levels, and suggests that the dual-stream architecture effectively leverages atom-level structural reasoning (through the primal stream) and bond-level interaction reasoning (through the line graph stream) at the appropriate stages of generation.

\end{document}